\documentclass[journal]{IEEEtran}
\relax
\usepackage{times}  %Required
\usepackage{helvet}  %Required
\usepackage{courier}  %Required
\usepackage{url}  %Required
\usepackage{graphicx}  %Required
\usepackage{url}
\usepackage{colortbl}
\usepackage {graphicx}
\usepackage{color}
\usepackage {subfigure}
\usepackage{setspace}
\usepackage{url}
\usepackage{colortbl}
\usepackage {graphicx}
\usepackage[final]{pdfpages}
\usepackage{color}
\usepackage {subfigure}

\usepackage{hhline}
\def\R{{\mathbb{R}}}

\usepackage[linesnumbered,ruled,vlined]{algorithm2e}
\usepackage{amsmath}
\usepackage{graphicx}
\usepackage{multirow}
\usepackage{subfigure}
\usepackage{wrapfig}
\usepackage{soul}
\usepackage[small,it]{caption}

\usepackage{subfigure} 
\usepackage{amsmath}
\usepackage{amsthm}
\usepackage{mathrsfs}
\usepackage{amsfonts}
\usepackage{multirow}
\usepackage{amssymb}
\usepackage{mathtools,xparse}
\usepackage{listings}
\usepackage{comment}
\usepackage{multirow}

\ifCLASSINFOpdf
\else
\fi
\begin{document}
\title{Segmentation-Aware Image Denoising without Knowing True Segmentation}

\author{Sicheng~Wang,~\IEEEmembership{Student Member,~IEEE,}
        Bihan~Wen,~\IEEEmembership{Member,~IEEE,} Junru~Wu,~\IEEEmembership{Student Member,~IEEE,} 
        ~Dacheng~Tao,~\IEEEmembership{Fellow,~IEEE,} and Zhangyang~Wang,~\IEEEmembership{Member,~IEEE}% <-this % stops a space
\thanks{S. Wang, J. Wu and Z. Wang are with the Department
of Computer Science and Engineering, Texas A\&M University, College Station,
TX, 77843 USA, e-mail: \{sharonwang, sandboxmaster, atlaswang\}@tamu.edu.}% <-this % stops a space
\thanks{B. Wen is with the School of Electrical \& Electronic Engineering, Nanyang Technological University, Singapore, e-mail: bihan.wen@ntu.edu.sg.}% <-this % stops a space
\thanks{D. Tao is with the School of Computer Science, the University of Sydney, NSW 2006 Australia, e-mail: dacheng.tao@sydney.edu.au.}}% <-this % stops a space
\maketitle

\begin{abstract}
Several recent works discussed application-driven image restoration neural networks, which are capable of not only removing noise in images but also preserving their semantic-aware details, making them suitable for various high-level computer vision tasks as the pre-processing step.
However, such approaches require extra annotations for their high-level vision tasks, in order to train the joint pipeline using hybrid losses. The availability of those annotations is yet often limited to a few image sets, potentially restricting the general applicability of these methods to denoising more unseen and unannotated images. Motivated by that, we propose a segmentation-aware image denoising model dubbed \textbf{U-SAID}, based on a novel \textbf{unsupervised} approach with a pixel-wise uncertainty loss. U-SAID does not need any ground-truth segmentation map, and thus can be applied to any image dataset. It generates denoised images with comparable or even better quality, and the denoised results show stronger robustness for subsequent semantic segmentation tasks, when compared to either its supervised counterpart or classical ``application-agnostic'' denoisers. Moreover, we demonstrate the superior \textbf{generalizability} of U-SAID in three-folds, by plugging its ``universal'' denoiser without fine-tuning: (1) 
denoising unseen types of images; (2) denoising as pre-processing for segmenting unseen noisy images; and (3) denoising for unseen high-level tasks, 
%as well as robustness for image segmentation, comparing to the popular ``application-agnostic'' image denoising methods. Furthermore, the proposed U-SAID demonstrated superior generalizability for (1) denoising unseen types of images, (2) segmentation of unseen noisy images, and (3) denoising for unseen high-level vision tasks, by plugging its ``universal'' denoiser without fine-tuning.
Extensive experiments demonstrate the effectiveness, robustness and generalizability of the proposed U-SAID over various popular image sets.
\end{abstract}

% \begin{IEEEkeywords}

% \end{IEEEkeywords}

% \fi
\IEEEpeerreviewmaketitle

\section{Introduction}
\IEEEPARstart{I}{mage} denoising aims to recover the underlying clean image signal from its noisy measurement. It has been traditionally treated as an independent signal recovery problem, focusing on either single-level fidelity (e.g., PSNR) or human perception quality of the recovery results. However, once high-level vision tasks are conducted on noisy images and such a separate image denoising step is typically applied as preprocessing, it will become suboptimal because of its unawareness of semantic information. A series of recent works \cite{liu2017image,cheng2017robust,li2017aod,fan2018segmentation,liu2018connecting,liu2019enhance} discussed \textit{application-driven image restoration models} that are capable of simultaneously removing noise and preserving semantic-aware details for certain high-level vision tasks. Those models achieve visually promising denoising results with richer details, in addition to better utility when supplied for high-level task pre-processing. 

However, a common drawback of them is their demand for \textit{extra annotations} for the high-level vision tasks, in order to train the joint pipeline with hybrid low-level and high-level supervisions. On one hand, such annotations (e.g., object bounding boxes, semantic segmentation maps) are often highly non-trivial to obtain for real images, therefore limiting current works to synthesizing noise on existing annotated clean datasets, to demonstrate the effectiveness of their methods. On the other hand, training with only one annotated dataset runs the risk of overly tying the resulting denoiser with the semantic information of this specific dataset, which causes a lack of universality and may show various artifacts due to overfitting, when applied to denoising other substantially different images.

This paper attempts to break the above hurdles of existing application-driven image restoration models. We propose a novel \textit{unsupervised segmentation-aware image denoising} (\textbf{U-SAID}) model, that enforces segmentation awareness and discriminative ability of denoisers, \textbf{without actually needing any segmentation groudtruth during training}. It is implemented by creating a novel loss term, that penalizes the \textit{pixel-wise uncertainty} of the denoised outputs for segmentation. Our contributions are in two-folds:
\begin{itemize}
\item On the \textit{low-level} vision side, to the best of our knowledge, U-SAID is the first unsupervised (or ``self-supervised'') application-driven image restoration model. In contrast to the existing peer work \cite{liu2017image}, U-SAID can be trained on any image datatset, without needing ground-truth (GT) segmentation maps. That greatly extends the applicability of U-SAID as a more ``universal'' denoiser, that can be applied to denoise images with few semantic annotations while being substantially different from natural images in existing segmentation datasets. 
Compared to standard ``application-agnostic'' denoisers such as \cite{zhang2017beyond}, U-SAID is observed to provide better visual details, that are also more favored under perception-driven metrics \cite{mittal2013making}. 

\item On the \textit{high-level} vision side, the U-SAID denoising network is shown to be robust and ``universal'' enough, when applied to denoising different noisy datasets, as well as when used towards boosting the segmentation task performance on unseen noisy datasets, thanks to its less semantic association with any dataset annotation. Furthermore, U-SAID trained with segmentation awareness generalizes well to unseen high-level vision tasks, and can be plugged into without fine-tuning, which reduces the training effort when applied to various high-level tasks.
\end{itemize}

Extensive experiments on various popular image sets demonstrating the outstanding effectiveness, robustness, and universality of the proposed approach. We advocate that our methodology is (almost) a \textit{free lunch} for image denoising, and has a plug-and-play nature to be incorporated with existing deep denoising models.

\section{Related Work}

Image denoising has been studied with intensive efforts for decades. Earliest methods refer to various image filters \cite{tomasi1998bilateral}. Later on, many model-based method with various priors have been introduced to this topic, in either spatial or transform domain, or their hybrid, such as spatial smoothness \cite{rudin1992nonlinear}, non-local patch similarity \cite{dabov2007image}, sparsity \cite{elad2006image,mairal2009non,octobos} and low-rankness \cite{gu2014weighted}. More recently, a number of deep learning models have demonstrated superior performance for image denoising~\cite{burger2012image,mao2016image,zhang2017beyond}. Despite their encouraging process, most existing denoising algorithms reconstruct images by minimizing the mean square error (MSE), which is well-known to be mis-aligned with human perception quality and often tends to over-smooth textures \cite{hore2010image}. Moreover, while image denoising algorithms are often needed as the pre-processing step for the acquired noisy visual data before subsequent high-level visual analytics, their impact on the semantic visual information was much less explored. 

Lately, a handful of works are devoted to closing the gap between the low-level (e.g., image denoising, as a representative) and high-level computer vision tasks. Such marriage leads to, not only better utility performance for high-level target tasks, but also the denoising outputs with richer visual details after receiving the extra semantic guidance from the high-level tasks, the latter being first revealed in \cite{johnson2016perceptual,wang2016studying}. \cite{liu2017image} presented a systematical study on the mutual influence between the low-level and high-level vision networks. The authors cascaded a fixed pre-trained semantic segmentation network after a denoising network, and tuned the entire pipeline with a joint loss function of MSE and segmentation loss. In that way, the authors showed the denoised images to have sharper edges and clearer textual details, as well as higher segmentation and classification accuracies when feeding such denoised images for those tasks. A similar effort was described in \cite{fan2018segmentation}, where a segmentation-aware deep fusion network was proposed to utilize the segmentation labels in MRI datasets to aid MRI compressive sensing recovery. \cite{li2017aod} considered a joint pipeline of image dehazing and object detection. \cite{shen2018deep} proposed to incorporate global semantic priors (e.g., eyes and mouths) as an input to deblur the highly structured face images. This field is now rapidly growing, with a few benchmarks launched recently \cite{li2019benchmarking,vidalmata2019bridging,li2019single,yuan2019ug}. 

Following \cite{liu2017image,fan2018segmentation}, we also adopt segmentation as our high-level task, because it can supply pixel-wise feedbacks and is thus considered to be more helpful for dense regression tasks. As pointed out by \cite{harley2017segmentation}, the availability of segmentation information can compromise the over-smoothening effects of CNNs across regions and increases their spatial precision. However, we would like to emphasize (again) that while \cite{liu2017image,harley2017segmentation,fan2018segmentation} all exploit GT segmentation maps as extra strong \textit{supervision} information during training,  we have only a weaker form of \textit{feedbacks} available from the segmentation task, due to the absence of its GT as extra information. Straightforwardly, our methodology is applicable when cascaded with other high-level tasks as well. 

Our work is also broadly related to training deep network with noisy or uncertain annotations \cite{veit2017learning,lu2017learning}. Especially for the segmentation task, existing supervised models require manually labeled segmentations for training. But pixel-based labeling for high-resolution images is often time-consuming and error-prone, causing incorrect pixel-wise annotations. Existing works often consider them as label noise \cite{reed2014training}. For example, \cite{li2017noise} proposed a noise-tolerant deep model for histopathological image segmentation, using the label-flip noise models proposed in \cite{sukhbaatar2014training}. However, those algorithms still need to be given segmentation maps (though inaccurate), and often demand more statistical estimations of the label noise.

%The proposed idea is also broadly related to unsupervised visual learning \cite{pathak2016context,doersch2017multi}. \cite{chen2018reblur2deblur} proposed to fine-tunes existing deep deblurring model in a self-supervised fashion by enforcing that the output, when blurred based on the optical flow between sub-equent frames, matches the input blurry image. In comparison, we use the segmentation map estimated from the input noisy image to regularize the denoising process, and can thus also be viewed as a self-supervised 
%- Not included, because we still use GT clean image to supervise

\section{The Proposed Model: U-SAID}

Our proposed unsupervised segmentation-aware image denoising (U-SAID) network follows the same cascade idea of the segmentation-guided denoising framework proposed by \cite{liu2017image}. We replace their self-designed U-Net denoiser with the classical deep denoiser DnCNN \cite{zhang2017beyond}, using the 20-layer blind color image denoising model referred to as CDnCNN-B\footnote{\url{https://github.com/cszn/DnCNN}}, since we favor more robustness to varying noise labels. Note that the choice of denoiser network should not affect much our obtained conclusions. Its loss $L_{MSE}$ is the reconstruction MSE between the denoised output and the clean image. 

\begin{figure*}[t]
  \centering
  \includegraphics[width=\textwidth]{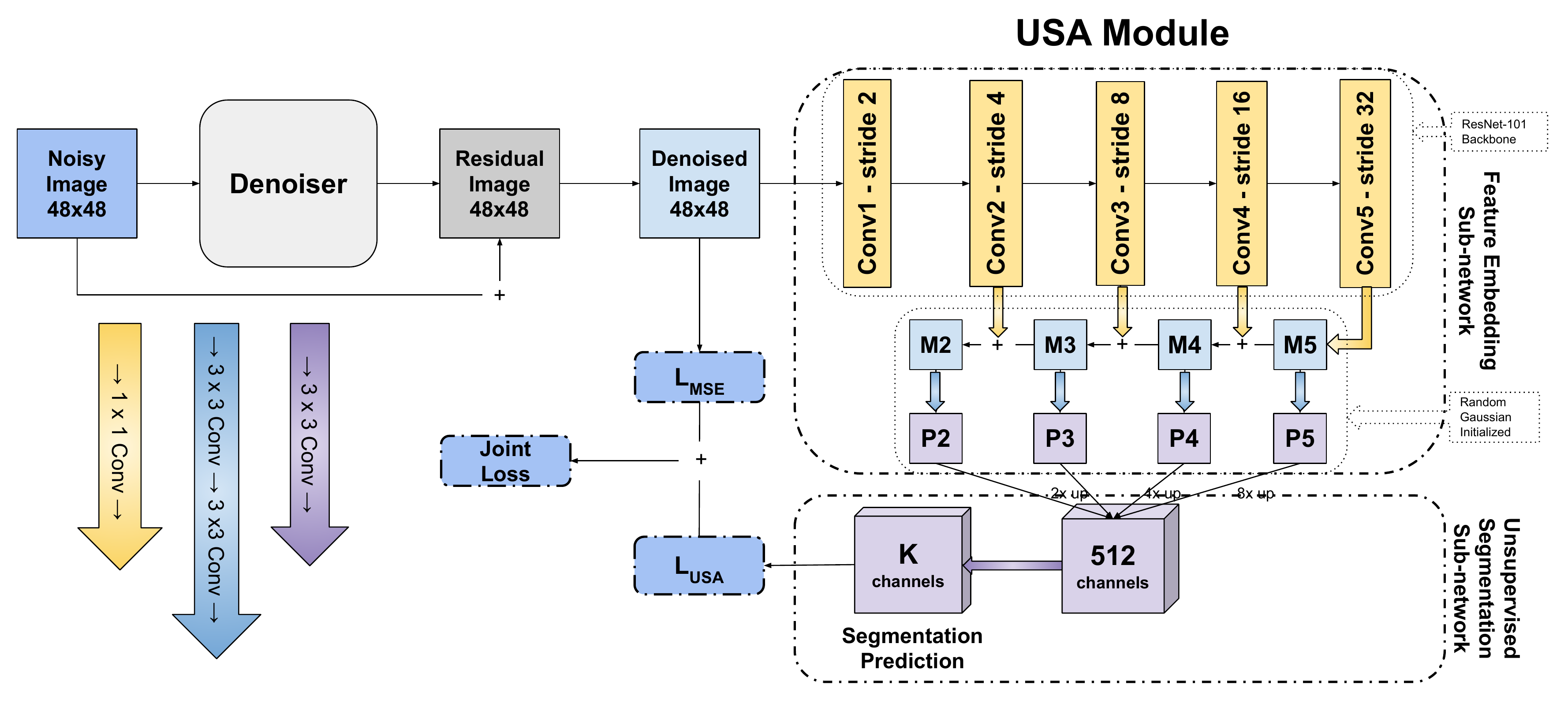}
  \caption{The architecture of the proposed U-SAID network}
%  \vspace{-2em}
    \label{flowchart}
\end{figure*}

The critical difference between U-SAID and existing works lies in the high-level component of the cascade. Unlike \cite{liu2017image,fan2018segmentation} that placed a pre-trained and fixed segmentation network with true segmentation labels given for training, we design a new \textit{unsupervised segmentation awareness} (\textbf{USA}) module, that requires no segmentation labels to train with. The network architecture is illustrated in Figure \ref{flowchart}.  

\subsection{Design of USA Module}

The USA module is composed of a feature embedding sub-network for transforming the input (denosied image) to the feature space, followed by an unsupervised segmentation sub-network that calculates the \textit{pixel-wise uncertainty} of semantic segmentation. 

\underline{For the feature embedding sub-network}, we used a Feature Pyramid Network (FPN) \cite{lin2017feature}, with a ResNet-101 backbone as the feature encoder. We used ImageNet-pretrained weights\footnote{\url{https://github.com/pytorch/vision/blob/master/torchvision/models/resnet.py}} for the backbone, and keep all default architecture details of FPN/ResNet-101 unchanged. 
During training, the ResNet-101 backbone is frozen as a fixed feature extractor, and the top-down feature pyramid part of FPN started with random Gaussian initializations and also keeped fixed. \textit{It is very important to notice that we have not used any image segmentation dataset to pre-train the feature embedding sub-network.}

\underline{For the unsupervised segmentation sub-network}, we assume the input image resolution to be $M \times N$ and contain at most $K$ different semantic classes. After FPN, we obtain 512 channels of feature maps $\in \R^{\frac{M}{4} \times \frac{N}{4}}$. We then apply K 3 $\times$ 3 convolutions to re-organize the output feature maps into $K$ channels, eventually leading to a (resized) $K$-class segmentation map.
%use nearest-neighbor interpolation to resize them into a 3-D output of size $K \times M \times N$, which resembles a $K$-class segmentation map for the input. 

%the true segmentation map to be in the same size as the output by nearest-neighbor interpolation.
%each of which then goes through nearest-neighbor interpolation to be of $M \times N$ resolution.
%followed by batch normalization and ReLU activation. We also applied spatial dropout \cite{tompson2015efficient} to alleviate overfitting. We then utilize 1 $\times$ 1 convolutions to 

Since the image segmentation task can be casted as pixel-wise classification, classical segmentation networks will adopt pixel-wise softmax loss function to generate a $K$-class probability vector $p_{i,j}$, for the $(i,j)$-th $\R^K$ vector ($i, j$ range from 1 to $M,N$, respectively), choosing the highest probability class and producing the final segmentation map $\in \R^{M \times N}$. However, since we have no GT pixel labels in the unsupervised case, we instead minimize the average entropy function of all predicted class vectors $p_{i,j}$, denoted as $L_{USA}$, to encourage confident predictions at all pixels:
\[
L_{USA} = \frac{1}{MN} \sum_{1\le i \le M, 1\le j \le N} - p_{i,j} \log p_{i,j}
\]
%Different from the feature embedding sub-network, 
% All layer-wise weights in the unsupervised segmentation sub-network need to be trained from random Gaussian initializations. Hence the entire USA module, except the ResNet-101 backbone with pre-trained ImageNet weights, is learned from scratch jointly with the denoiser. That is different from all existing works using fixed high-level networks \cite{liu2017image,fan2018segmentation}, as well as perceptual loss. 
All layer-wise weights in the unsupervised segmentation sub-network are random Gaussian initialized, and the ResNet-101 backbone uses the pre-trained ImageNet weights. Similar to \cite{liu2017image,fan2018segmentation}, we use a fixed high-level network, but we do not include the perceptual loss in training the network. 

% SegNet \cite{badrinarayanan2017segnet}
% Focal loss \cite{lin2018focal}

\subsection{Training Strategy}
We train the cascade of denoising network and USA module in an end-to-end manner, while fixing the weights in the feature embedding sub-network of the latter. The overall loss for U-SAID is: $L_{MSE} + \gamma L_{USA}$, with the default $\gamma = 1$ unless otherwise specified. The training dataset for U-SAID could be any image set and is unnecessary to have segmentation annotations, overcoming the limitations in \cite{liu2017image,fan2018segmentation}. That said, we need an estimate of segmentation class numbers $K$ to construct $L_{USA}$: an ablation study of estimated $K$ will follow. 

We use the Adam solver to train both the denoiser part and the USA module. The batch size is 16. The input patches are set to be 48 $\times$ 48 pixels (patches are randomly sampled from images with a stride of 1). The initial learning rate is set as 1e-3 for all learnable parts of U-SAID, using a multi-step learning decay strategy, i.e. dividing the learning rate by 10 at epoch 10, 40 and 80, respectively.
%and is divided by 10 after every ** iterations. 
The training is terminated after 100 epochs. 

%%Therefore it is better to merely train a new denoising network for each level of image corruption while keeping unchanged the well-trained network for high-level vision task, rather than retrain the whole cascaded network.

\subsection{Why It Works?}
% A noteworthy feature of U-SAID is its (partially) trainable high-level network, together with the denoiser. Without strong label supervision, one may wonder why it can regularize the denoiser training effectively, since it merely trains several convolutional layers from random weights, to regress ResNet-101 ImageNet features into some unknown map, that is only required to be low-entropy pixel-wise. In fact, if the learnable part has large enough capacity, one may expect to be able to find parameters that can fit with any given pixel-wise map (low-entropy or not), that conveys little semantical information (e.g., random maps).
A noteworthy feature of U-SAID is frozen high-level network, together with the denoiser. 
Without strong label supervision, one may wonder why it can regularize the denoiser training effectively, since it is high level features include the random initialization keep fixed, and the ResNet-101 ImageNet features can still be regressed into some unknown map, that is only required to be low-entropy pixel-wise. In fact, if the network itself holds large enough capacity, one may expect to be able to find parameters that can fit with any given pixel-wise map (low-entropy or not), that conveys little semantical information (e.g., random maps).

That might have reminded the \textit{deep image prior} proposed in \cite{UlyanovVL17}: the authors first trained a convolutional network from random scratch, to regress from a random vector to a given corrupted image, and then used the trained network as a regularization. Since no aspect of the network is pre-trained from data, such deep image prior is effectively handcrafted and was shown to work well for various image restoration tasks. The authors attributed the success to the convolutional architecture itself, that appeared to possess high noise impedance. In our case, the ImageNet features are thought as highly relevant to image semantics. Therefore, we make the similar hypothesis with the authors of \cite{UlyanovVL17}: although the parametrization may regress to any random unstructured label map, it does so very reluctantly. 

%\textcolor{red}{[TO DO add the convergence plot]}
\begin{figure}[h!]
\includegraphics[width=\linewidth]{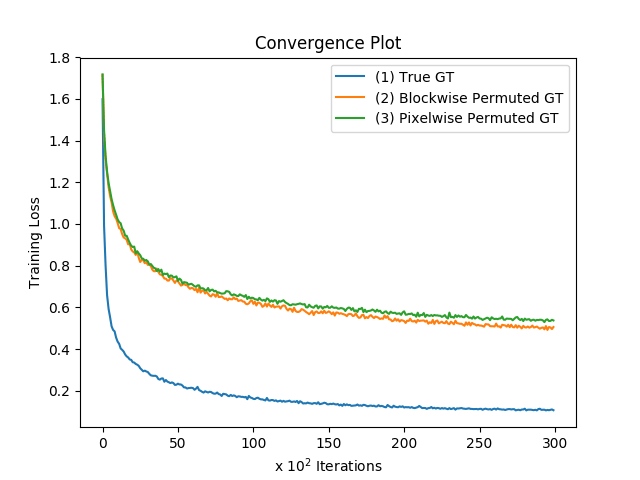}
  \caption{Convergence Plot}
  \label{fig:convergence_plot}

\end{figure}

To verify our hypothesis, we conduct a simple proof-of-concept experiment inspired by \cite{zhang2016understanding}. In the USA module, we replace $L_{USA}$ with a standard pixel-wise softmax loss, having ResNet-101 fixed with ImageNet weights and other parts initialized randomly. We then use PASCAL VOC 2012 training set to train this modified USA module, in a supervised way, but with three different choices for the supervision: 1) the GT segmentation maps; 2) evenly cutting each GT map into 4 sub-images, and randomly permuting their locations; 3) randomly permuting all pixel locations in each GT map. Notice that if we compute $L_{USA}$ values for the three target maps, they should be the same. 
%the loss values of $L_{USA}$ should be the same for these three maps. \textcolor{red}{(may need to be modified here)}
%We then train a segmentation network over those feature embeddings, with the same structure as the U-SAID unsupervised segmentation sub-network, but this time in a supervised way with pixel-wise softmax loss. 

We show in Figure \ref{fig:convergence_plot} the value of training loss, as a function of the gradient descent iterations for three supervisions. Apparently, the network can converge much faster to GT maps; the more GT maps were permuted, the more convergence ``inertia'' we observe. In other words, the network descends much more quickly towards semantically meaningful maps, and resists ``bad'' solutions with fewer semantics, although their entropies might have been the same.

\section{Experiments}

\subsection{Denoising Study on PASCAL-VOC}

The U-SAID denoiser takes RGB images as input and outputs the reconstructed images. We choose the PASCAL-VOC 2012 training set, and add i.i.d. Gaussian noise with zero mean and standard deviation $\sigma$ to synthesize the noisy input image during training. Our testing set is generated similarly by adding noise on the PASCAL-VOC 2012 validation set. Since we used CDnCNN-B as the backbone denoiser, we focus on the challenging blind denoising scenario, by setting the Gaussian noise standard deviations $\sigma$ to uniformly range between [0, 55] for the training set, creating a ``one-for-all'' denoiser that can be simply evaluated at different testing sets with various $\sigma$s. 
%We crop original images into 48 $\times$ 48 patches for training, with a batch size of 16. 
The PASCAL-VOC 2012 sets have 20 classes of interested objects, plus a background class, leading to $K$ = 21 unless otherwise specified. 

We compare U-SAID with the 
%classical CBM3D \cite{dabov2007image} and the 
original CDnCNN-B (re-trained on our training set) \cite{zhang2017beyond}, which requires no segmentation information at all. We further create another denoisr following the same idea of \cite{liu2017image}: cascading CDnCNN-B with the supervised segmentation network (i.e., replacing $L_{USA}$ with a standard pixel-wise softmax loss), with all other training protocols and initialization the same as U-SAID. We call it \textit{supervised segmentation-aware image denoising} (\textbf{S-SAID}), and train it with the hybrid MSE-segmentation loss (the two losses are weighted equally), using the ground-truth segmentation maps available on the PASCAL training set. 
\textbf{Note that S-SAID is the only method that exploits ``true'' segmentation information}, making it a natural baseline for U-SAID to show the effect of such \textit{extra information}. We do not include other denoising methods such as \cite{dabov2007image,burger2012image,gu2014weighted} because: 1) their average performance was shown to be worse than CDnCNN; and 2) most of them are not designed for the blind denoising scenario, thus hard to make fair comparisons. 
We have exhaustively tuned the hyper-parameters (learning rates, etc.) for CDnCNN-B and S-SAID, to ensure the optimal performance of either baseline. 

\begin{table}%[th!]
	\begin{center}
	%	\fontsize{8}{10pt}\selectfont
		\caption{The average image denoising performance comparison on PASCAL-VOC 2012 validation set, with $\sigma$ = 15, 25, 35. \textcolor{red}{Red} is the best and \textcolor{blue}{blue} is the second best results (the same hereinafter)}
		\label{table:pascal_denoising}		
% 		\vspace{-2mm}
		\begin{tabular}{|@{\hskip 1mm}c@{\hskip 1mm}|@{\hskip 1mm}c@{\hskip 1mm}|c|@{\hskip 1mm}c@{\hskip 1mm}|@{\hskip 1mm}c@{\hskip 1mm}|}		% 5 columns
			\hline
			\multicolumn{2}{|c|}{}  & CDnCNN-B & S-SAID & U-SAID \\   [-0.3ex] % reduce the space between two rows
			\hline 
			\hline
			\multirow{3}{*}{$\sigma$=15} %& PSNR & 24.94 & 32.58 & \textcolor{red}{32.96} & 32.08 & \textcolor{blue}{32.88}  & 32.74 \\ [-0.3ex]
			%\hline	
			& PSNR (dB)  &  \textcolor{red}{33.56} & 33.40 & \textcolor{blue}{33.50}\\ [-0.3ex]
			%\hline	
			& SSIM   & \textcolor{red}{0.9159} & 0.9136 & \textcolor{blue}{0.9153} \\
            & NIQE  & 4.3290 & \textcolor{blue}{4.0782} & \textcolor{red}{4.0049} \\ 
			\hline
						\multirow{3}{*}{$\sigma$=25} %& PSNR & 24.94 & 32.58 & \textcolor{red}{32.96} & 32.08 & \textcolor{blue}{32.88}  & 32.74 \\ [-0.3ex]
			%\hline	
			%\hline	
			& PSNR (dB)  & \textcolor{red}{31.18} & 31.01 & \textcolor{blue}{31.13}\\
            & SSIM   & \textcolor{red}{0.8725} & 0.8698 & \textcolor{blue}{0.8724}\\ 
            & NIQE  & 4.2247 & \textcolor{red}{3.8508} & \textcolor{blue}{3.8975}\\ [-0.3ex]
			\hline
			\multirow{3}{*}{$\sigma$=35} %& PSNR & 24.94 & 32.58 & \textcolor{red}{32.96} & 32.08 & \textcolor{blue}{32.88}  & 32.74 \\ [-0.3ex]
			%\hline	
			& PSNR (dB)  & \textcolor{red}{29.65}  & 29.47 & \textcolor{blue}{29.59} \\ [-0.3ex]
			%\hline	
			& SSIM  & \textcolor{blue}{0.8344} & 0.8312 & \textcolor{red}{0.8347}  \\
            & NIQE  & 4.1022 & \textcolor{red}{3.6679} & \textcolor{blue}{3.7612}\\ 
			\hline
		\end{tabular}
		
	\end{center}
%	\vspace{-2mm}
%	\vspace{-5mm}
\end{table}

The typical metric used for image denoising is PSNR, which has been shown to correlate poorly with human assessment of visual quality \cite{huynh2008scope}. On the other hand, in the metric of PSNR, a model trained by minimizing MSE on the image domain should always outperform a model trained by minimizing a hybrid weighted loss. Therefore, we emphasize that the goal of our following experiments is not to pursue the highest PSNR, but to quantitatively demonstrate the different behaviors between models with and without segmentation awareness. 

Table \ref{table:pascal_denoising} reports the denoising performance in terms of PNSR, SSIM and Naturalness Image Quality Evaluator (NIQE) \cite{mittal2013making}. The last one is a well-known no-reference image quality score to indicate the perceived ``naturalness'' of an image: a smaller score indicates better perceptual quality. Our observations from Table \ref{table:pascal_denoising} are summarized as below:
\begin{itemize}
\item Since CDnCNN-B is optimized towards the MSE loss, it is not surprising that it consistently achieves the best PSNR results among all. However, U-SAID is able to achieve \textit{only marginally inferior} PSNR/SSIMs to CDnCNN-B, which usually surpass S-SAID. 
\item The two methods with segmentation awareness (U-SAID and S-SAID) are significantly more favored by NIQE, showing a large margin over CDnCNN-B (e.g., nearly 0.4 at $\sigma = 25$). That testifies the benefits of considering high-level tasks for denoising. 
\item While not exploiting the true segmentation maps during training as S-SAID did, the performance of U-SAID is almost as competitive as S-SAID under the NIQE metric. In other words, \textit{we did not lose much without using the true segmentation as supervision}.
\end{itemize}

%Since CDnCNN-B is optimized with the MSE loss, it is not surprising that that it consistently achieves the best PSNR results among all. U-SAID is able to achieve slightly higher PSNR and SSIM values than S-SAID. However, we observe that the results from the two methods with segmentation awareness (U-SAID and S-SAID) are much more favored by NIQE, showing a large margin over CDnCNN-B (e.g., 0.5 at $\sigma = 25$). It is encouraging to observe that while not exploiting the true segmentation maps during training as S-SAID did, U-SAID is still capable to perform competitively under perceptual metrics. 

% \begin{table}[h]
% \small
% \begin{center}
% \begin{tabular}{|c|c|c|c|c|c|}
% \hline
% $K$ & 10 & 15 & 21 (default) & 25  & 40 \\ \hline
% PSNR & 31.00& 31.06 & \textcolor{red}{31.13} & 30.99 & 30.98\\ \hline
% SSIM & 0.8688& 0.8707 & \textcolor{red}{0.8724} & 0.8668  & 0.8673\\ \hline
% NIQE & 3.9878 & \textcolor{red}{3.8320} & 3.8975 & 4.1139 & 3.9746\\ \hline
% \end{tabular}
% \end{center}
% \caption{Ablation study of varying $K$ in U-SAID training. }
% \label{table:pascal_K}
% 	\vspace{-5mm}
% \end{table}

\begin{table}[h]
\small
\begin{center}
\setlength{\tabcolsep}{2.5pt}
\begin{tabular}{|c|c|c|c|c|c|c|c|}

\hline
$K$ & 10 & 15 & 20 & 21 (default) & 22 & 25  & 40 \\ \hline \hline
NIQE & 3.9878 &  \textcolor{red}{3.8320} & 4.0783 & 3.8975 & 3.8455& 4.1139 & 3.9746\\ \hline 
PSNR & 31.00  & 31.06 & 30.99 & \textcolor{red}{31.13} & 31.01 & 30.99 & 30.98\\ \hline
% SSIM & 0.8688& 0.8707 & \textcolor{red}{0.8724} & 0.8668  & 0.8673\\ \hline
\end{tabular}
\end{center}
\vspace{-0.5em}
\caption{Ablation study of varying $K$ in U-SAID training. }
\label{table:pascal_K}
% 	\vspace{-3mm}
\end{table}

\paragraph{Ablation Study on ``Unsupervised Segmentation''} In training U-SAID above, we have used the ``true'' class number $K$ = 21. It is then to our curiosity that: is this ground-truth value really best for training denoisers? Or, if the class number information cannot be accurately inferred when tackling general images, how much the denoising performance might be affected? 

We hereby present an ablation study, by training several U-SAID models with different $K$ values (all else remain unchanged), and compare their denoising performance on the testing set, as displayed in Table \ref{table:pascal_K}. It is encouraging to observe that, the U-SAID denoising performance (PSNR and SSIM) consistently increase as $K$ grows from smaller values (10, 15) towards the true value (21), and thens gradually decreases as $K$ get further larger. The NIQE values show the similar first-go-up-then down trend, except the peak slightly shifted to 15. That acts as a side evidence that rather than learning a semantically blind discriminator, the USA module indeed picks up the semantic class information and benefits from the correct $K$ estimate. On the other hand, the variations of denoising performance w.r.t $K$ are mild and smooth, showing certain robustness to inaccurate $K$s too.

\paragraph{More Comparison to Relevant Methods}
% As reviewer 1 correctly pointed out: “As far as comparison with CDnCNN-B and created S-SAID is considered, it seems to be justifiable ... and most of them are not designed for the blind denoising scenario. This [is] hard to make fair comparisons.”

To solidify our results, we include more off-the-shelf denoising methods for comparison. We performed these experiments on Kodak dataset with three test sigmas 15, 25 and 35. A detailed comparison for each method we use is shown in \ref{table:method}. However, all methods we mentioned previously, i.e. CDnCNN-B, S-SAID and U-SAID, are blind to the noise level, the competing methods are non-blind. Therefore, we created two settings to simulate blind denoising: 
\begin{itemize}
    \item Applying the median sigma as denoising input ($\sigma = 25$);
    \item Assuming the oracle sigma is known in denoising
\end{itemize} The second setting is apparently unfair to our blind model. Even so, we demonstrate the results in \ref{table:compare}, from which U-SAID constantly yields the best performance.

\begin{table}[ht]
\begin{center}
\setlength{\tabcolsep}{2.5pt}
\caption{Comparison of different methods. The three categories (columns) verify if the methods i) are using deep learning, ii) are semantic-aware denoising methods, and iii) require extra segmentation annotation. }
\label{table:method}
\begin{tabular}{|c|c|c|c|}
\hline
        & Deep  & Semantic & Segmentation  \\ 
        & Learning   & -Aware & Annotation \\ \hline\hline
U-SAID  & \checkmark & \checkmark &  \\ \hline
S-SAID  & \checkmark & \checkmark     & \checkmark  \\
\hline\hline
% S-SAID  & &  label required    &  \\ \hline
CDnCNN-B  & \checkmark &                 &  \\ \hline
MLP \cite{burger2012image}     & \checkmark &   &  \\ \hline\hline
MC-WNNM \cite{xu2017multi} &            &  &  \\ \hline
CBM3D \cite{dabov2007color}  & &  &   \\ \hline
\end{tabular}
\end{center}
\vspace{-2mm}
    % \label{table:method}
\end{table}

\begin{table}[h]%[th!]
	\begin{center}
		\fontsize{9}{10pt}\selectfont
		\setlength{\tabcolsep}{5pt}
% 		\label{table:compare}		
% 		\vspace{-2mm}
% 		\begin{tabular}{|@{\hskip 1mm}c@{\hskip 1mm}|@{\hskip 1mm}c@{\hskip 1mm}|@{\hskip 1mm}c@{\hskip 1mm}|c|@{\hskip 1mm}c@{\hskip 1mm}|}		% 5 columns
% 		\begin{tabular}{|@{\hskip 1mm}c@{\hskip 1mm}|@{\hskip 1mm}c@{\hskip 1mm}|@{\hskip 1mm}c@{\hskip 1mm}|c|@{\hskip 1mm}|}
		\begin{tabular}{|c|c|c|c|}
		   
% 		\begin{|c|c|c|c|c|}
        
			\hline
% 			\hline
% 			\multicolumn{2}{|@{\hskip 27.5mm}|}{}  & $\sigma$=15 & $\sigma$=25 & $\sigma$=35 \\   [-0.3ex] % reduce the space between two rows
            \multicolumn{4}{|c|}{Setting I} \\ 
            \hline \hline
            \multicolumn{1}{|c|}{}  & $\sigma$=15 & $\sigma$=25 & $\sigma$=35 \\  
			\hline 
% 			\hline
		
			\multirow{1}{*}{MLP \cite{burger2012image}} %& PSNR & 24.94 & 32.58 & 
			 & 4.3924/ 29.83  & 3.0205/ 30.09 & 6.5367/ 23.50\\ 
			\multirow{1}{*}{MC-WNNM \cite{xu2017multi}} %& PSNR & 24.94 & 32.58 & 
			 & 5.6334 / 31.04  & 3.6731/ 31.35 & 8.6496/ 21.53 \\ 
			\multirow{1}{*}{CBM3D \cite{dabov2007color}} %& PSNR & 24.94 & 32.58 & 
			 & 3.7707/ 32.60  & 2.6152/ 31.81 & 6.7044/ 25.29 \\ 
			\hline 
			\multicolumn{4}{|c|}{Setting II} \\ 
            \hline 
            % \hline
			\multicolumn{1}{|c|}{}  & $\sigma$=15 & $\sigma$=25 & $\sigma$=35 \\  
			\hline 
			\hline
			
% 			Non- &
			\multirow{1}{*}{MLP \cite{burger2012image}} %& PSNR & 24.94 & 32.58 & 
			 & 4.675/ 29.11  & 3.008/ 30.09 & 3.070/ 28.67\\ 
% 			Blind &
			\multirow{1}{*}{MC-WNNM \cite{xu2017multi}} %& PSNR & 24.94 & 32.58 & 
			 & 3.302/ 33.94  & 3.673/ 31.35 & 4.039/ 29.70 \\ 
% 			\hline 
			\multirow{1}{*}{CBM3D \cite{dabov2007color}} %& PSNR & 24.94 & 32.58 & 
			 & 2.6360/ 34.40  & 2.6620/ 31.81 & 2.6786/ 30.04 \\ 
			\hline 
			
% 			\multirow{3}{*}{Blind} &
%     		\multirow{1}{*}{CDnCNN-B} %& PSNR & 24.94 & 32.58 & 
% 			 &  2.757/ \textcolor{red}{34.75} & 2.849/ \textcolor{red}{32.27} & 2.975/ \textcolor{red}{30.69} \\ 
% 			&
%     		\multirow{1}{*}{S-SAID} %& PSNR & 24.94 & 32.58 & 
% 			 &  \textcolor{blue}{2.629}/ 34.57 & \textcolor{red}{2.601}/ 32.07 & \textcolor{red}{2.562}/ 30.48 \\
% 		     &
%     		\multirow{1}{*}{U-SAID} %& PSNR & 24.94 & 32.58 & 
% 			 &  \textcolor{red}{2.569}/ \textcolor{blue}{34.62} & \textcolor{blue}{2.636}/ \textcolor{blue}{32.17} & \textcolor{blue}{2.669}/ \textcolor{blue}{30.50} \\ 
% 			\hline
		\end{tabular}
	\vspace{-3mm}		
	\end{center}
	\caption{The average Image denoising performance comparison in NIQE/ PSNR on the Kodak dataset, with noise $\sigma$ = 15, 25, 35, respectively. }
	\label{table:compare}

\end{table}

%But somewhat surprisingly, even better results can be achieved at larger $K$ (**) than $K$ = 21, before dropping as $K$ increases further. 

%Why this ``mismatch''? We notice that on PASCAL-VOC, the ``background'' is conventionally considered as one big class. But in fact, it contains many finer-grain classes and is thus semantically more complicated/richer than the other 20 object-of-interests classes. Therefore, without strong supervision from pixel labels (like in S-SAID), U-SAID does not necessarily have to treat all possible background objects as one class, and may instead prefer ``over-segmenting'' them into more finer-grain classes. However, although the default $K$ = 21 might not always be the best for denoising, we still adopt it by default for simplicity. 

% As a side product of denoising with U-SAID, we can simultaneously obtain an ``unsupervised segmentation map'' for each image, by rounding each pixel in the USA module output to its nearest one-hot vector (although we cannot name their semantic categories). Figure ** displays an image from PASCAL-VOC 2012 \underline{training} set, its group truth segmentation map, and the unsupervised segmentation map estimated from U-SAID output\footnote{There is no semantic class information directly available from U-SAID output. We manually choose the ``correct'' colors for the segmented foreground and background pixel clutters by maximum overlap, while each other pixel region uses a random color.}.
% \textcolor{red} {TO DO Figure of segmentation map visualization}. 

\subsection{Segmentation Study on PASCAL-VOC}

\begin{table}[h]
\begin{center}
\begin{tabular}{|c|c|c|c|c|}
\hline
& noisy  & CDnCNN-B & S-SAID & U-SAID \\ \hline
$\sigma$=15  & 0.4227&  0.4238 & \textcolor{red}{0.4349}  & \textcolor{blue}{0.4336} \\ \hline
$\sigma$=25  &0.4007 & 0.4003 & \textcolor{red}{0.4084} & \textcolor{blue}{0.4047} \\ \hline
$\sigma$=35  &0.3667 & 0.3724 & \textcolor{red}{0.3802} & \textcolor{blue}{0.3785} \\ \hline
\end{tabular}
\end{center}
\vspace{-2mm}
% if CBM3D is not competitive in Table 1, then no need to discuss any more
\caption{Segmentation results (mIoU) after denoising noisy image
inputs, averaged over Pascal VOC 2012 validation dataset.}
%\vspace{-0.5em}
		\label{table:seg}
\end{table}

\begin{figure*}[t!]
\begin{center}
\begin{tabular}{c@{\hskip 0.5mm}c@{\hskip 0.5mm}c@{\hskip 0.5mm}c@{\hskip 0.5mm}c@{\hskip 0.5mm}c}
\includegraphics[width=0.195\linewidth, trim=0 0 0 36, clip]{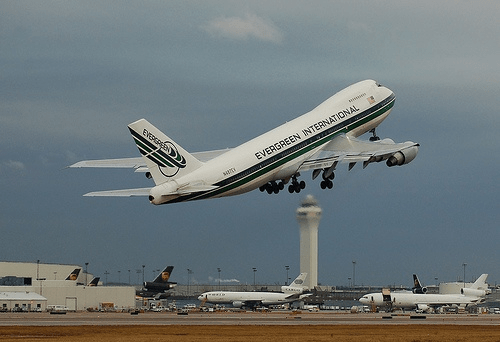}
&
\includegraphics[width=0.195\linewidth, trim=0 0 0 36, clip]{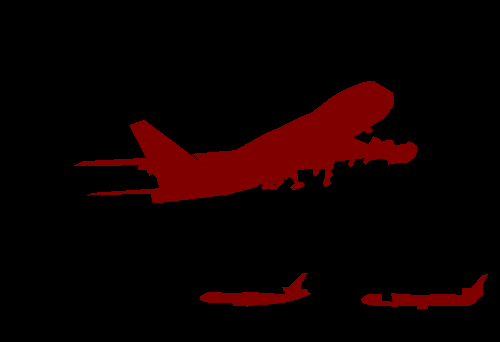} &
\includegraphics[width=0.195\linewidth, trim=0 0 0 36, clip]{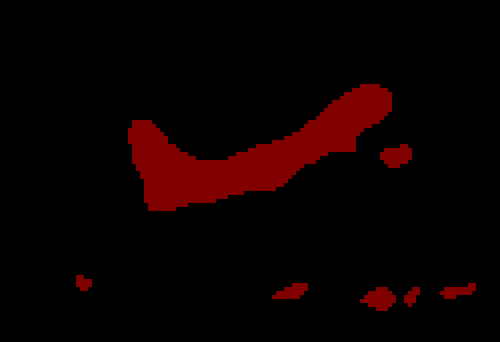} & 
\includegraphics[width=0.195\linewidth, trim=0 0 0 36, clip]{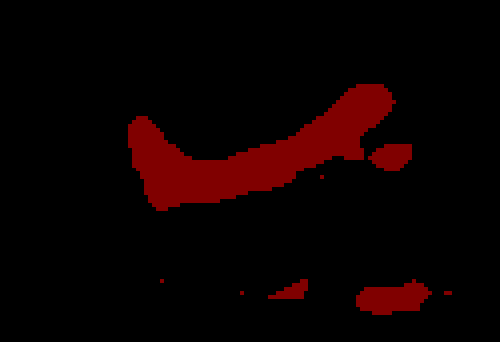} &
%{\small (a)} & {\small (c)} & {\small (e)} & {\small (g)} \\
\includegraphics[width=0.195\linewidth, trim=0 0 0 36, clip]{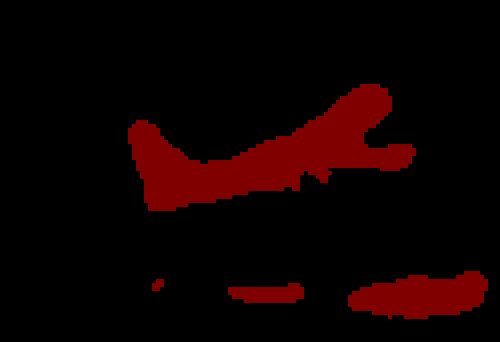} &
\\
 &  &  IOU: 0.7866 & IOU: 0.7909 & IOU: 0.7872 \\
\includegraphics[width=0.195\linewidth, trim=0 0 0 36, clip]{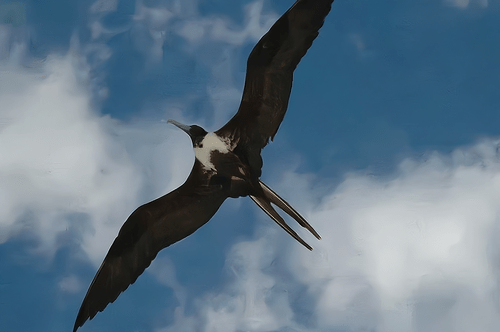}
&
\includegraphics[width=0.195\linewidth, trim=0 0 0 36, clip]{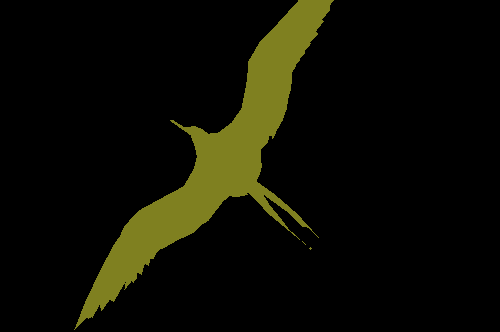} &
\includegraphics[width=0.195\linewidth, trim=0 0 0 36, clip]{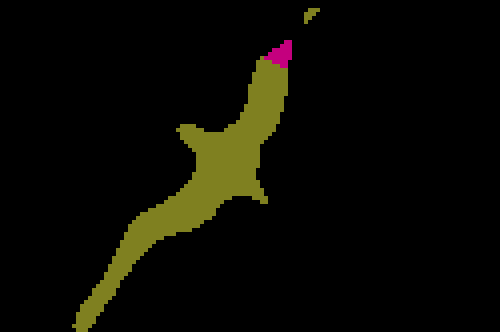} & 
\includegraphics[width=0.195\linewidth, trim=0 0 0 36, clip]{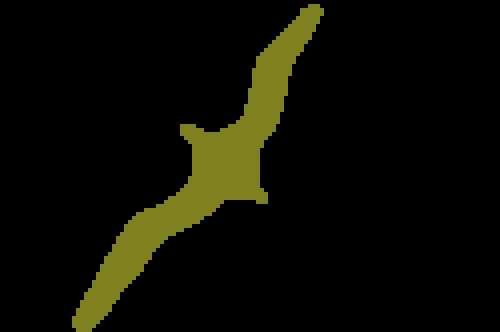} &
%{\small (a)} & {\small (c)} & {\small (e)} & {\small (g)} \\
\includegraphics[width=0.195\linewidth, trim=0 0 0 36, clip]{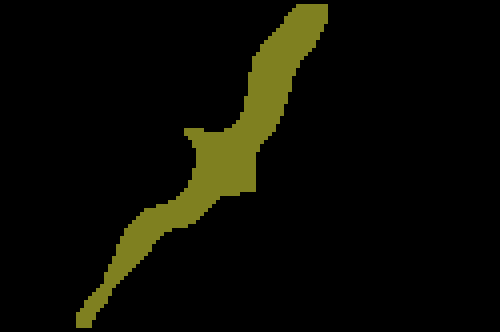} &
\\
 & &  IOU: 0.5432 & IOU: 0.8827 & IOU: 0.8720 \\
\includegraphics[width=0.195\linewidth, trim=0 0 0 36, clip]{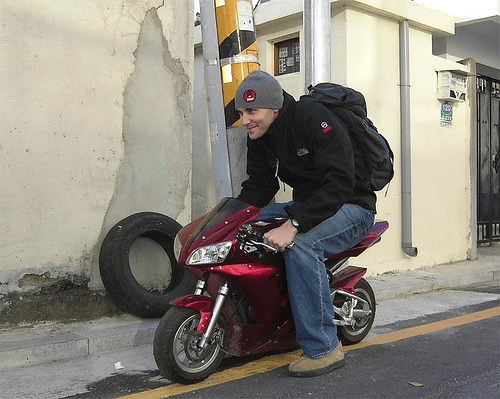} &
\includegraphics[width=0.195\linewidth, trim=0 0 0 36, clip]{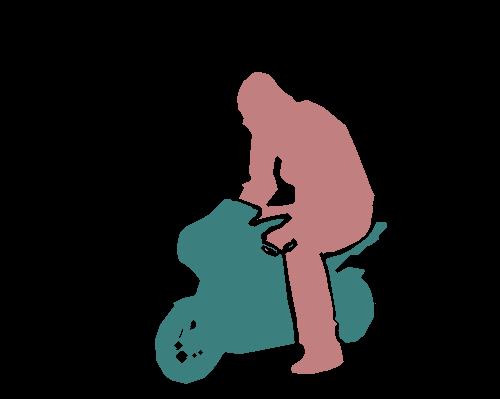} &
\includegraphics[width=0.195\linewidth, trim=0 0 0 36, clip]{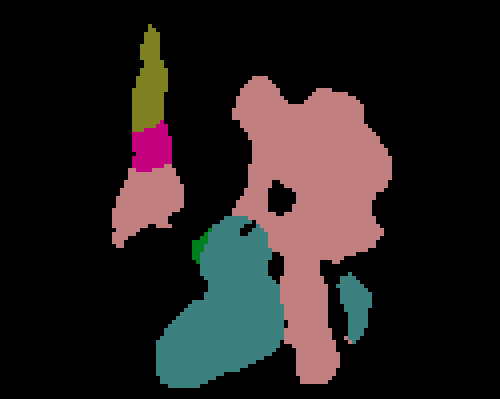} & 
\includegraphics[width=0.195\linewidth, trim=0 0 0 36, clip]{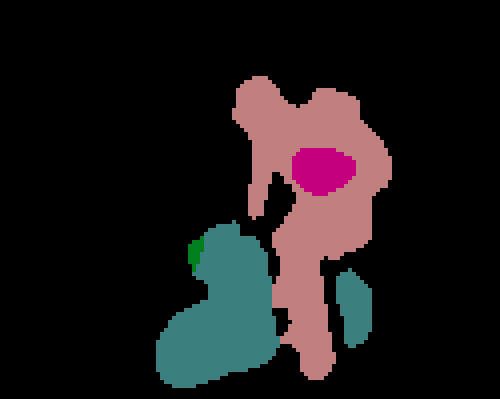} &
%{\small (a)} & {\small (c)} & {\small (e)} & {\small (g)} \\
\includegraphics[width=0.195\linewidth, trim=0 0 0 36, clip]{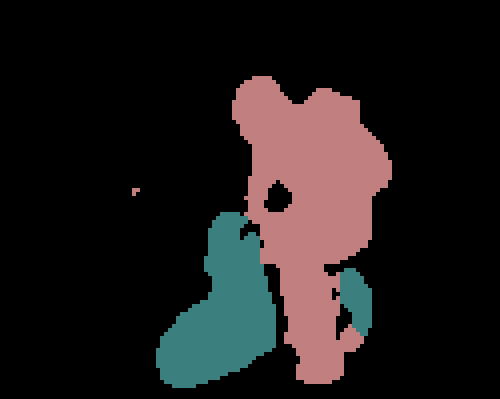} &
\\
 & &  IOU: 0.5432 & IOU: 0.8827 & IOU: 0.8720 \\
(a) Original Image & (b) True Segmentation & (c) C-DnCNNB & (d) S-SAID & (e) U-SAID \\

\end{tabular}
\end{center}
\vspace{-2mm}
\caption{Visualized semantic segmentation examples from Pascal VOC 2012 validation set. The first row is added with noise of $\sigma$ = 15, the second row $\sigma$ = 25 and the third row $\sigma$ = 35. Columns (a) - (b) are the ground truth images and true segmentation maps; (c) -(e) are the results by applying the pre-trained segmentation model on the denoised images using (c) C-DnCNNB; (d) S-SAID; and (e) U-SAID.  
%Their corresponding segmentation label maps are shown below. The zoom-in region is displayed in the red box.
}
\label{fig:seg1}
%\vspace{-2mm}
\end{figure*}

 We next investigate the effectiveness of denoising as a pro-processing step for the semantic segmentation over noisy images, which follows the setting in \cite{liu2017image}. We first pass the noisy images in the PASCAL-VOC testing set through each of the three learned denoisers (CDnCNN-B, S-SAID, and U-SAID). We then apply a FPN \textit{pre-trained on the clean PASCAL-VOC 2012 training set}, on the denoised testing sets, and evaluate the segmentation performance in terms of mean intersection-over-union (mIOU). 
 %Note this setting is critically different from the previous analysis on unsupervised segmentation. 
 
 As compared in Table \ref{fig:seg1}, when we apply the CDnCNN-B denoiser without considering high-level semantics, it easily fails to achieve high segmentation accuracy due to the artifacts introduced during denoising (even though those artifacts might not be reflected by PSNR or SSIM). With their segmentation awareness, both S-SAID and U-SAID have led to remarkably higher mIOUs. Most impressively, U-SAID is comparable to S-SAID, provided that \textit{the former has never seen true segmentation information on this dataset (training set)}, while the latter does. Figure \ref{fig:seg1} has visually confirmed the impact of denoisers on the segmentation performance. 

%\textcolor{red} {TO DO may need change figure after full mIOU out}. 

\subsection{Generalizability Study: Data, Semantics, and Task}

In this section, we define and compare three aspects of general usability, which were often overlooked in previous research of learning-based denoisers: 
\begin{itemize}
\item \textbf{Data Generalizability}: whether a denoiser trained on one dataset can be applicable to restoring another.
\item \textbf{Semantic Generalizability}: whether a denoiser trained on one dataset can be effective in preserving semantics, as the preprocessing step for applying semantic segmentation over another noisy dataset (with unseen classes).
\item \textbf{Task Generalizability}: whether a denoiser trained with segmentation awareness can also be effective as preprocessing for other high-level tasks over noisy images.
%\vspace{-1mm}
\end{itemize}
Throughout the whole section below, all three denoisers used are the same models trained on PASCAL-VOC 2012 above. \textbf{There is no re-training involved}. 

\textbf{Our hypothesis} is that since U-SAID is not trained with any annotation on the original training set, it may less likely overfit the training set's semantics than U-SAID, while still preserving discriminative features, and hence could generalize better to various unseen data, semantics and tasks.
%\vspace{-1mm}

\begin{figure*}[t!]
\begin{center}
\begin{tabular}{c@{\hskip 0.5mm}c@{\hskip 0.5mm}c@{\hskip 0.5mm}c@{\hskip 0.5mm}c@{\hskip 0.5mm}c@{\hskip 0.5mm}c}
\includegraphics[width=0.245\linewidth, trim=0 0 0 36, clip]{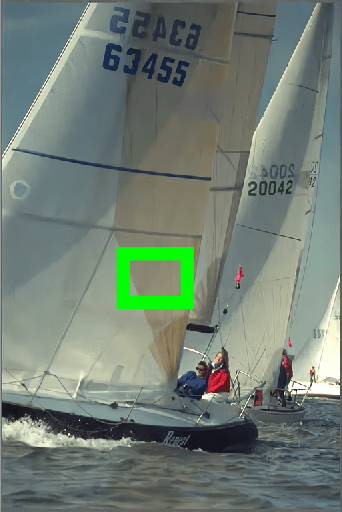} &
\includegraphics[width=0.245\linewidth, trim=0 0 0 36, clip]{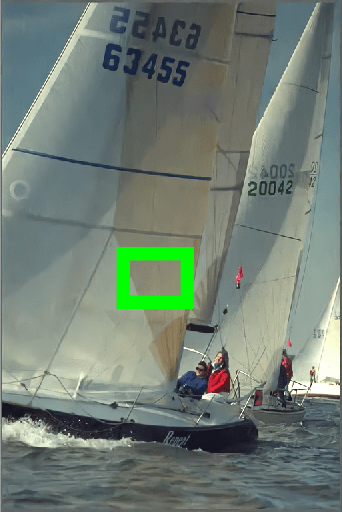} & 
\includegraphics[width=0.245\linewidth, trim=0 0 0 36, clip]{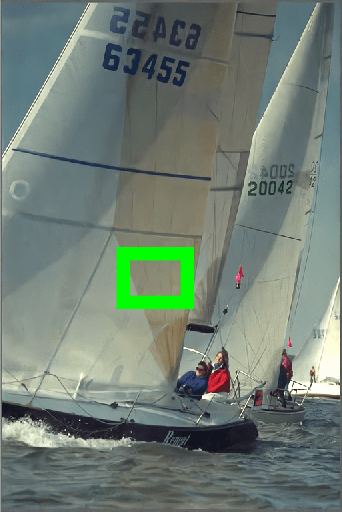} &
%{\small (a)} & {\small (c)} & {\small (e)} & {\small (g)} \\
\includegraphics[width=0.245\linewidth, trim=0 0 0 36, clip]{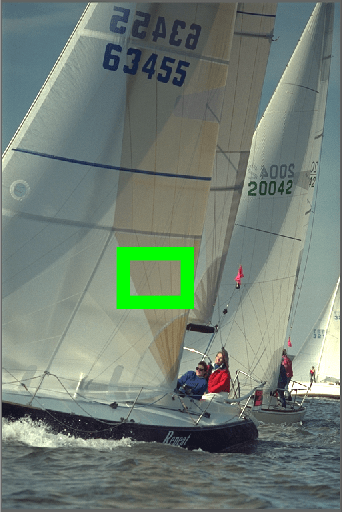} &
\\
\includegraphics[width=0.245\linewidth, trim=0 0 0 36, clip]{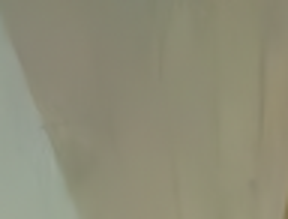} & 
\includegraphics[width=0.245\linewidth, trim=0 0 0 36, clip]{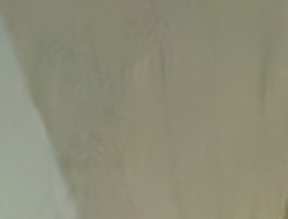} &
%{\small (a)} & {\small (c)} & {\small (e)} & {\small (g)} \\
\includegraphics[width=0.245\linewidth, trim=0 0 0 36, clip]{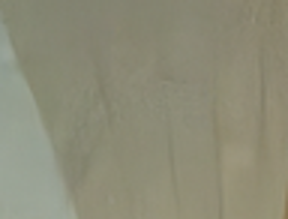} &
\includegraphics[width=0.245\linewidth, trim=0 0 0 36, clip]{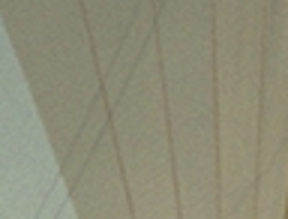} &
\\
{\small (PSNR = 34.31, NIQE = 3.01)} &  {\small(PSNR = 34.02, NIQE = 2.76)}  &  {\small(PSNR = 34.27, NIQE = 2.71)} & $\sigma = 25$ \\
(a) CDnCNN-B &  (b) S-SAID  & (c) U-SAID & (d) Ground Truth \\
\end{tabular}
\end{center}
\vspace{-1.5mm}
\caption{Visual comparison on one Kodak image. We show the full images (top) and zoom-in regions (bottom) of the ground truth as well as three denoised images by CDnCNN-B, S-SAID and U-SAID, at $\sigma$ = 25 (Best viewed on high-resolution color display, lower NIQE is better).  
%Their corresponding segmentation label maps are shown below. The zoom-in region is displayed in the red box.
}
\label{fig:kodak}
\vspace{-2mm}
\end{figure*}

\begin{figure*}[h!]
\begin{center}
\begin{tabular}{c@{\hskip 0.5mm}c@{\hskip 0.5mm}c@{\hskip 0.5mm}c@{\hskip 0.5mm}c@{\hskip 0.5mm}c@{\hskip 0.5mm}c}
% CDnCNN-B &  S-SAID  & U-SAID & Ground Truth \\
\includegraphics[width=0.245\linewidth, trim=0 0 0 36, clip]{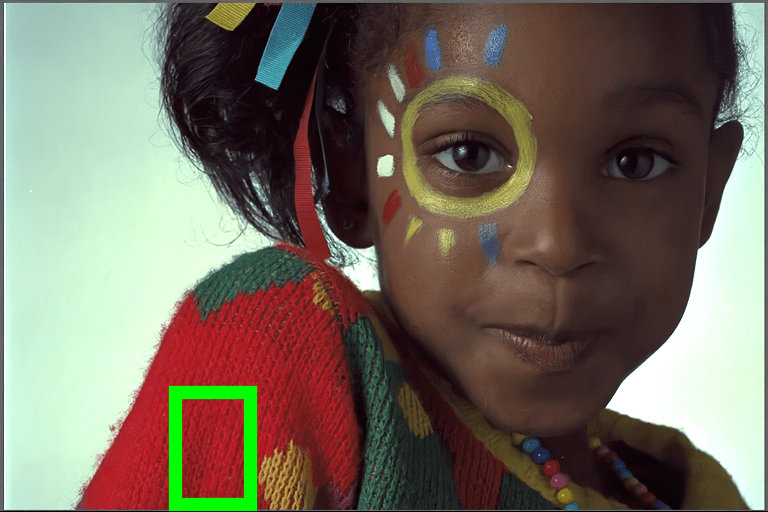} &
\includegraphics[width=0.245\linewidth, trim=0 0 0 36, clip]{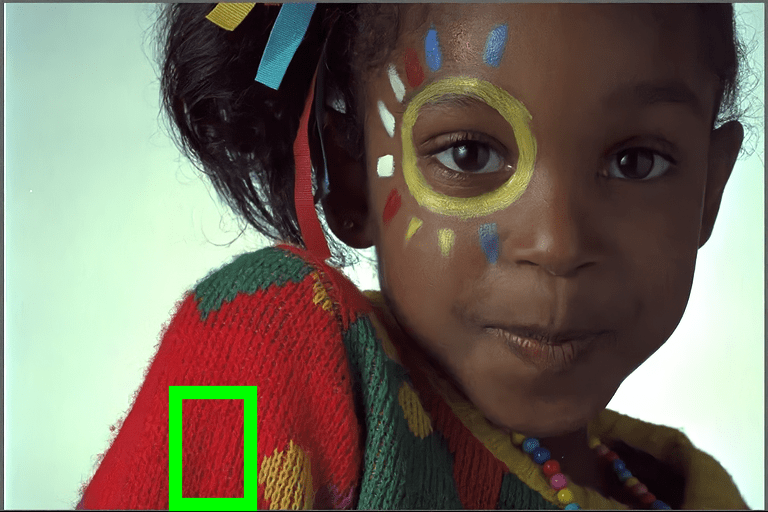} & 
\includegraphics[width=0.245\linewidth, trim=0 0 0 36, clip]{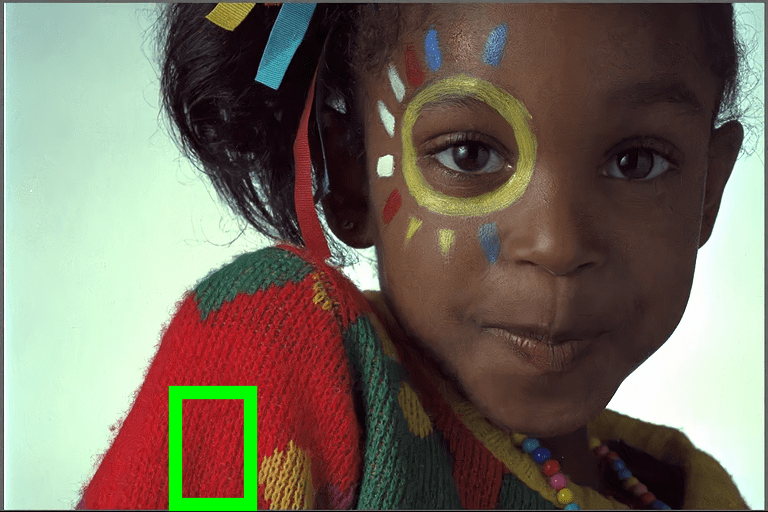} &
\includegraphics[width=0.245\linewidth, trim=0 0 0 36, clip]{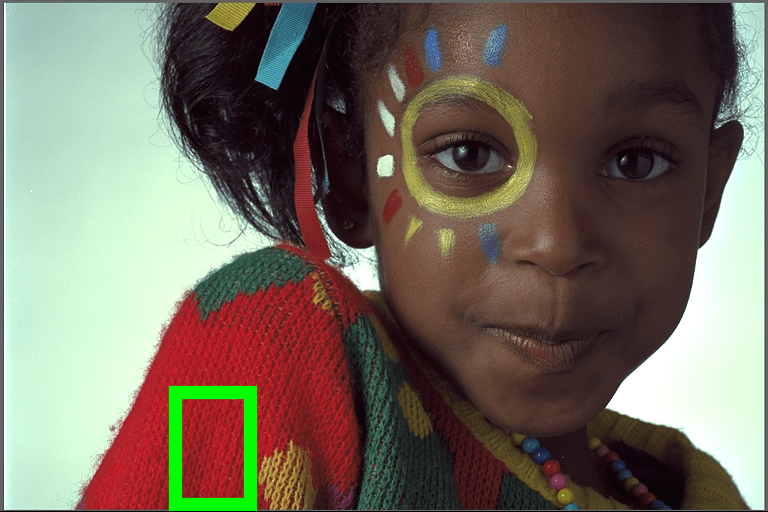} &
\\
\includegraphics[width=0.245\linewidth, trim=0 0 0 36, clip]{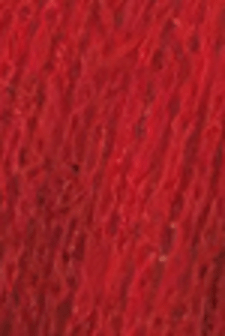} & 
\includegraphics[width=0.245\linewidth, trim=0 0 0 36, clip]{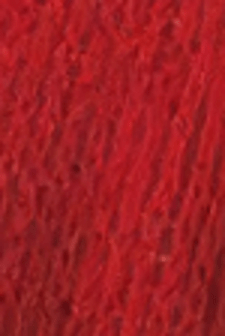} &
\includegraphics[width=0.245\linewidth, trim=0 0 0 36, clip]{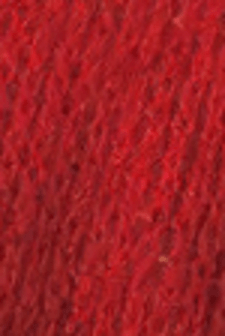} &
\includegraphics[width=0.245\linewidth, trim=0 0 0 36, clip]{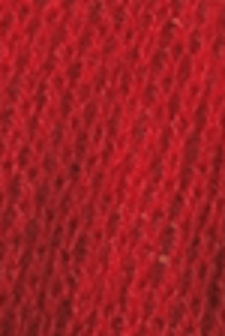} &
\\
% CDnCNN-B &  S-SAID  & U-SAID & Ground Truth \\
{\small (PSNR =35.52 , NIQE =3.0163)} &  {\small (PSNR =35.31 , NIQE =3.0701)}  & {\small (PSNR = 35.45, NIQE = 2.9791)} & $\sigma = 15$ \\

\includegraphics[width=0.245\linewidth, trim=0 0 0 36, clip]{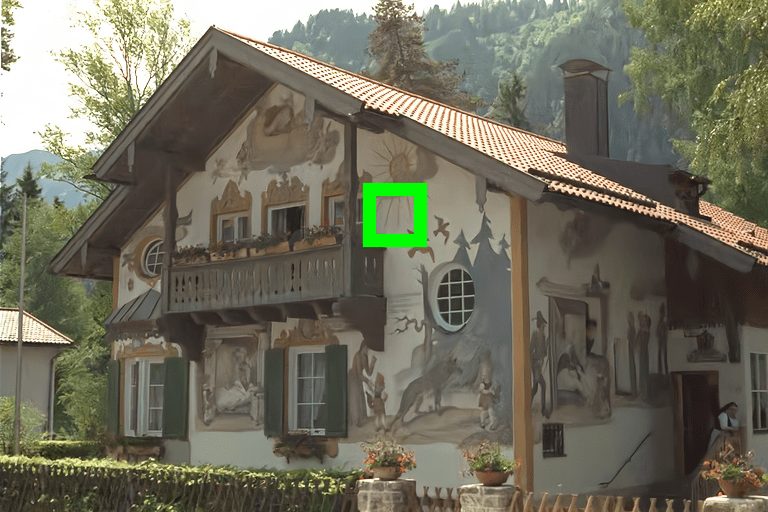} &
\includegraphics[width=0.245\linewidth, trim=0 0 0 36, clip]{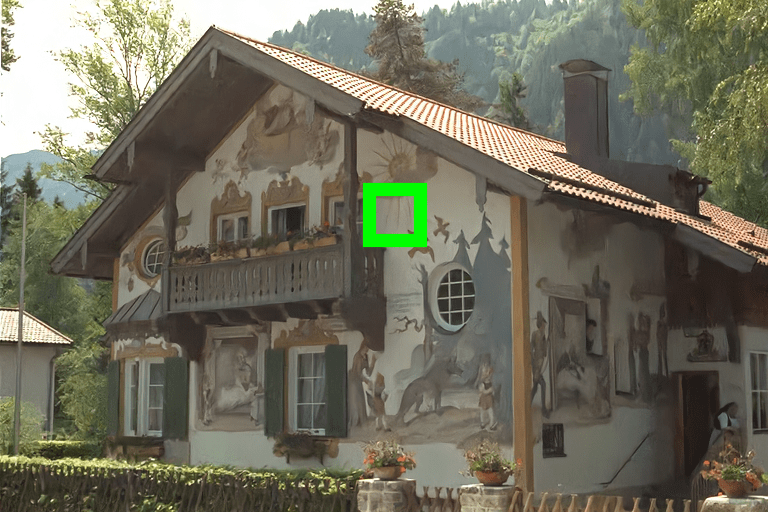} & 
\includegraphics[width=0.245\linewidth, trim=0 0 0 36, clip]{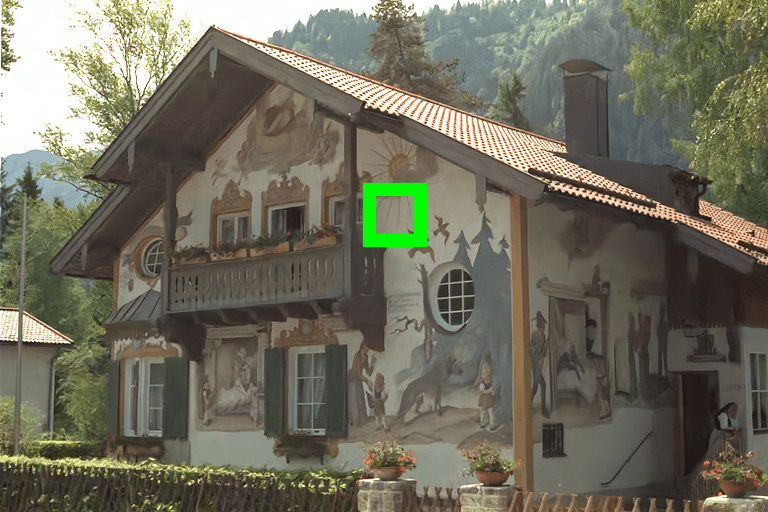} &
\includegraphics[width=0.245\linewidth, trim=0 0 0 36, clip]{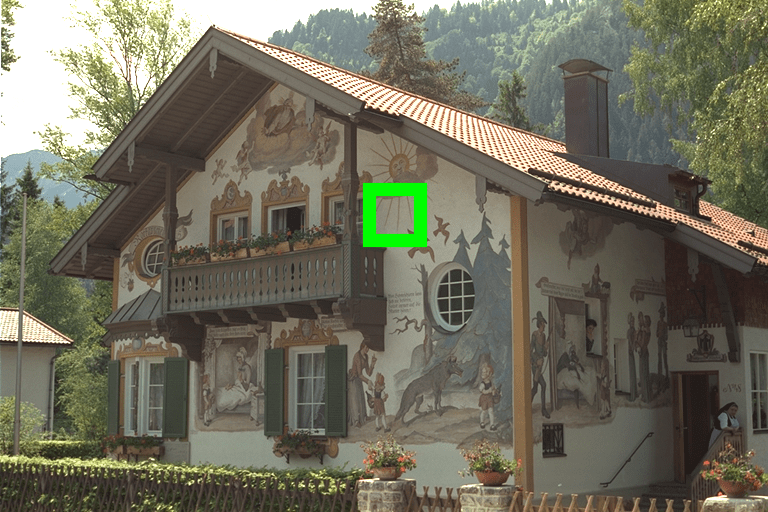} &
\\
\includegraphics[width=0.245\linewidth, trim=0 0 0 36, clip]{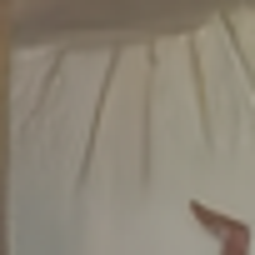} & 
\includegraphics[width=0.245\linewidth, trim=0 0 0 36, clip]{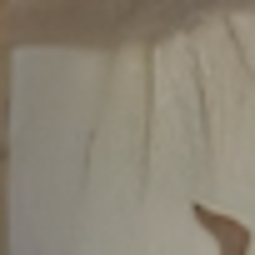} &
\includegraphics[width=0.245\linewidth, trim=0 0 0 36, clip]{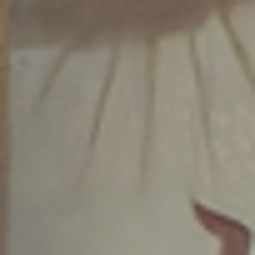} &
\includegraphics[width=0.245\linewidth, trim=0 0 0 36, clip]{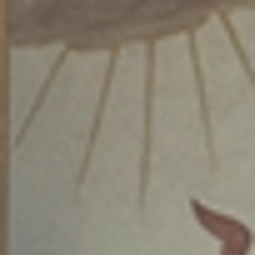} &
\\
% CDnCNN-B &  S-SAID  & U-SAID & Ground Truth \\
{\small (PSNR = 30.71, NIQE = 2.6303)} &  {\small (PSNR =30.49 , NIQE =1.9944)}  & {\small (PSNR =30.66 , NIQE = 2.3407)} & $\sigma = 25$ \\

\includegraphics[width=0.245\linewidth, trim=0 0 0 36, clip]{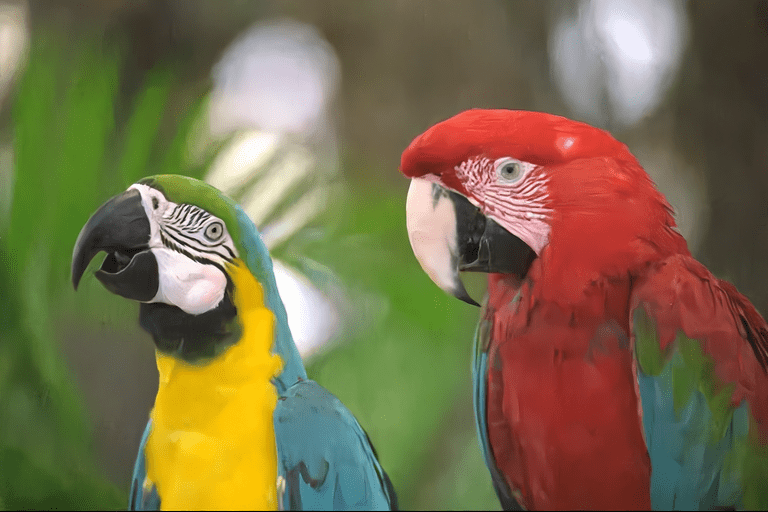} &
\includegraphics[width=0.245\linewidth, trim=0 0 0 36, clip]{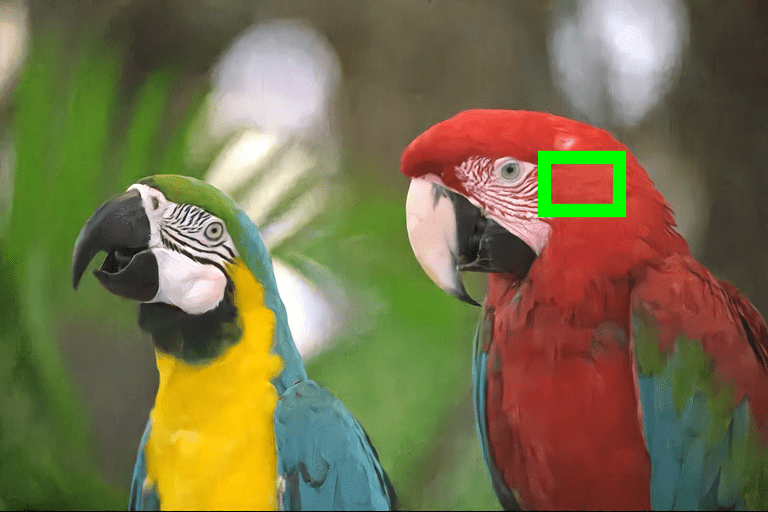} & 
\includegraphics[width=0.245\linewidth, trim=0 0 0 36, clip]{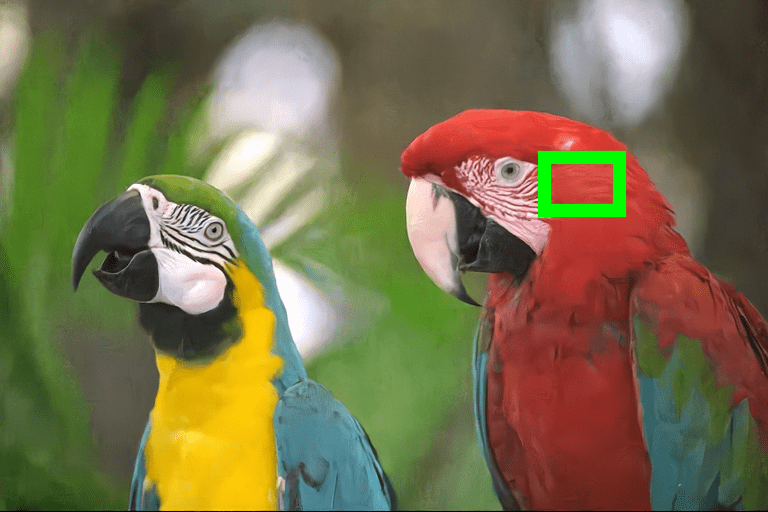} &
\includegraphics[width=0.245\linewidth, trim=0 0 0 36, clip]{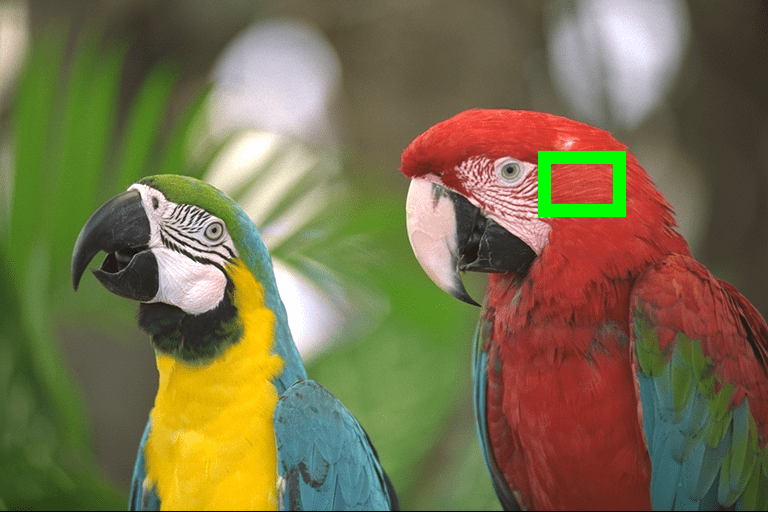} &
\\
\includegraphics[width=0.245\linewidth, trim=0 0 0 36, clip]{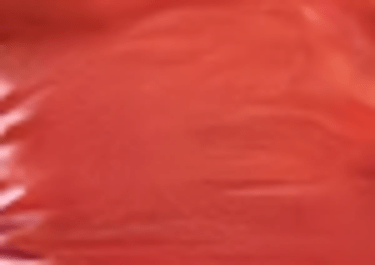} & 
\includegraphics[width=0.245\linewidth, trim=0 0 0 36, clip]{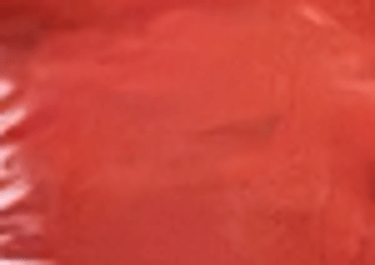} &
\includegraphics[width=0.245\linewidth, trim=0 0 0 36, clip]{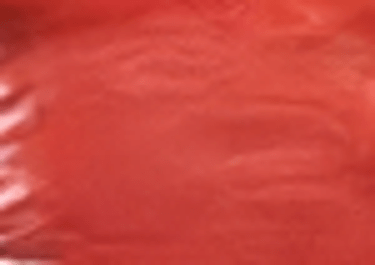} &
\includegraphics[width=0.245\linewidth, trim=0 0 0 36, clip]{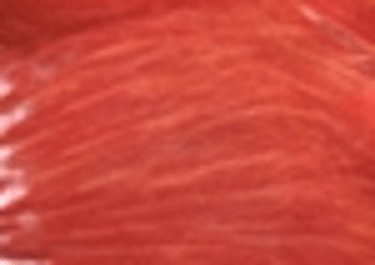} &
\\
% CDnCNN-B &  S-SAID  & U-SAID & Ground Truth \\
{\small (PSNR =33.75 , NIQE =3.6207)} &  {\small (PSNR =33.41 , NIQE = 3.0099)}  & {\small (PSNR = 33.63, NIQE = 3.0905)} & $\sigma = 35$ \\
(a) CDnCNN-B &  (b) S-SAID  & (c) U-SAID & (d) Ground Truth \\
% \vspace{-2mm}

\end{tabular}
\end{center}
\vspace{-1.5mm}
\caption{More denoised visualizations from Kodak data set by CDnCNN, S-SAID and U-SAID under three different noise level (lower NIQE indicates better visual quality). 
}
\label{fig:more}
\end{figure*}

% \vspace{-2mm}
\paragraph{Denoising Unseen Noisy Datasets}

We evaluate thee denoising performance over the widely used Kodak dataset\footnote{\url{http://r0k.us/graphics/kodak/}}, consisting of 24 color images. Table \ref{table:kodak_denoising} reports the quantitative results, which show strong consistency across all three noise levels: CDnCNN-B achieves the highest PSNR and SSIM values, while S-SAID performs the best in terms of NIQE. Interestingly, U-SAID seems to be the \textit{``balanced''} solution in terms of data generalizability: it tends to obtain very close PSNR and SSIM values compared to CDnCNN-B, while producing comparable or even better NIQE values to S-SAID (especially at smaller $\sigma$s).
We further observe that U-SAID is usually able to preserve sharper edges and textures than CDnCNN-B, sometimes even better than S-SAID. Figure \ref{fig:kodak} displays a group of examples, where U-SAID finds clear advantages in preserving local fine details on the sail. Please refering more visualizations to \ref{fig:more}.

\begin{table}%[th!]
	\begin{center}
	%	\fontsize{8}{10pt}\selectfont
		\caption{The average Image denoising performance comparison on the Kodak dataset, with noise $\sigma$ = 15, 25, 35, respectively. }
		\label{table:kodak_denoising}		
		\vspace{-2mm}
		\begin{tabular}{|@{\hskip 1mm}c@{\hskip 1mm}|@{\hskip 1mm}c@{\hskip 1mm}|c|@{\hskip 1mm}c@{\hskip 1mm}|@{\hskip 1mm}c@{\hskip 1mm}|}		% 5 columns
			\hline
			\multicolumn{2}{|c|}{}  & CDnCNN-B & S-SAID & U-SAID \\   [-0.3ex] % reduce the space between two rows
			\hline 
			\hline
			\multirow{3}{*}{$\sigma$=15} %& PSNR & 24.94 & 32.58 & \textcolor{red}{32.96} & 32.08 & \textcolor{blue}{32.88}  & 32.74 \\ [-0.3ex]
			%\hline	
			& PSNR & \textcolor{red}{34.75}  & 34.57 & \textcolor{blue}{34.62}\\ [-0.3ex]
			%\hline	
			& SSIM & \textcolor{red}{0.9242} & 0.9217 & \textcolor{blue}{0.9222}  \\
            & NIQE & 2.7570  & \textcolor{blue}{2.6288} & \textcolor{red}{2.5690}\\ 
			\hline
            			\multirow{3}{*}{$\sigma$=25} %& PSNR & 24.94 & 32.58 & \textcolor{red}{32.96} & 32.08 & \textcolor{blue}{32.88}  & 32.74 \\ [-0.3ex]
			%\hline	
			& PSNR &  \textcolor{red}{32.27} & 32.07 & \textcolor{blue}{32.17} \\ [-0.3ex]
			%\hline	
			& SSIM & \textcolor{red}{0.8812} & 0.8770 & \textcolor{blue}{0.8790}  \\
            & NIQE &  2.8493 & \textcolor{red}{2.6006} & \textcolor{blue}{2.6355} \\ 
			\hline
            			\multirow{3}{*}{$\sigma$=35} %& PSNR & 24.94 & 32.58 & \textcolor{red}{32.96} & 32.08 & \textcolor{blue}{32.88}  & 32.74 \\ [-0.3ex]
			%\hline	
			& PSNR &  \textcolor{red}{30.69} & 30.48 & \textcolor{blue}{30.50} \\ [-0.3ex]
			%\hline	
			& SSIM & \textcolor{red}{0.8418} & 0.8366 & \textcolor{blue}{0.8395}  \\
            & NIQE &  2.9753 & \textcolor{red}{2.5619} & \textcolor{blue}{2.6687} \\ 
			\hline
		\end{tabular}
	\vspace{-3mm}		
	\end{center}

\end{table}

% \vspace{-2mm}
\paragraph{Denoising for Unseen Dataset Segmentation} We choose two recently released real-world datasets, whose class categories are substantially different from PASCAL VOC: i) The ISIC 2018 dataset \cite{codella2018skin}\footnote{\url{https://challenge2018.isic-archive.com}}. We choose the validation set of Task 1: Lesion Segmentation, whose goal is to predict lesion segmentation boundaries from dermoscopic lesion images; ii) The DeepGlobe dataset\footnote{\url{http://deepglobe.org}}. We choose the validation set of Track 3: Land Cover Classification, whose goal is to predict a pixel-level mask of land cover types (urban, agriculture, rangeland, forest, water, barren, and unknow) from satellite images. 

\begin{table}[h]
%\small
\begin{center}
\begin{tabular}{|c|c|c|c|c|}
\hline
& noisy & CDnCNN-B & S-SAID & U-SAID \\ 
\hline \hline
ISIC 2018 & 0.8061 & 0.8076 &  \textcolor{blue}{0.8084} & \textcolor{red}{0.8095}   \\ \hline
DeepGlobe & 0.1309& \textcolor{blue}{0.4260} & 0.4198 & \textcolor{red}{0.4263}  \\ \hline
\end{tabular}
\end{center}
	\vspace{-2mm}
% if CBM3D is not competitive in Table 1, then no need to discuss any more
\caption{Segmentation results (mIoU) after denoising noisy image
inputs, on ISIC 2018 and DeepGlobe validation sets, respectively.}
		\label{table:seg_transfer}
%\vspace{-1em}
\end{table}

We add $\sigma$ = 25 noise to both validation sets, to create unseen testing sets for the trained denoisers. For either denoised validation set, we apply a pyramid scene parsing network (PSPNet) \cite{zhao2017pyramid}, that is pre-trained on the original clean training set. Table \ref{table:seg_transfer} reports the generalization effects of three denoisers when serving as preprocessing for segmenting unseen noisy datasets: U-SAID performs the best on both datasets, again verifying the benefits of segmentation awareness (that comes ``for free'' with no knowledge of true segmentation on any dataset). What is noteworthy, while we observe in the PASCAL-VOC segmentation experiment that the fully-supervised S-SAID is always superior to the segmentation-unaware CDnCNN-B, it is no longer always the case when applied to unseen datasets of different semantic categories: even CDnCNN-B is able to outperform S-SAID on DeepGlobe. Our hypothesis is that, the full supervision of S-SAID might cause its certain overfitting with PASCAL-VOC object categories. Trained in the unsupervised fashion but still equipped with segmentation awareness, U-SAID is not closely tied with original class semantics on the training set, and might thus generalize better to extracting and preserving semantics from new categories.

% \vspace{-2mm}
\paragraph{Denoising for Unseen High-Level Tasks}

We now investigate if the segmentation-aware image denoising can also enhance other high-level vision applications, and choose classification and detection as two representative examples. While also listing PSNR and SSIM, we primarily focus on comparing their utility metrics (i.e., accuracy and mAP). 

For classification, We choose the challenging CIFAR-100 dataset and add $\sigma$ = 25 noise to its validation set. We then pass it through three denoisers, followed by a ResNet-110 classification model, pre-trained on the clean CIFAR-100 training set. As seen from Table \ref{table:task_transfer}, while U-SAID is second best in terms of both PSNR and SSIM (marginally inferior to CDnCNN-B), it demonstrates a notable boost in terms of both top-1 and top-5 accuracies, with a good margin compared to CDnCNN-B and S-SAID. While S-SAID also outperforms CDnCNN-B in improving classification, U-SAID proves to have even better generalizablity here.

\begin{table}%[th!]
	\begin{center}
%		\fontsize{8}{10pt}\selectfont
		\caption{Classification results after denoising noisy image inputs ($\sigma$ = 25) from CIFAR-100.}
		\label{table:task_transfer}		
		\vspace{-2mm}
		\begin{tabular}{|@{\hskip 1mm}c@{\hskip 1mm}|c|@{\hskip 1mm}c@{\hskip 1mm}|@{\hskip 1mm}c@{\hskip 1mm}|@{\hskip 1mm}c@{\hskip 1mm}|}		% 5 columns
			\hline
			\multicolumn{1}{|c|}{} & noisy & CDnCNN-B & S-SAID & U-SAID \\   [-0.3ex] % reduce the space between two rows
			\hline 
			\hline
			%& PSNR & 24.94 & 32.58 & \textcolor{red}{32.96} & 32.08 & \textcolor{blue}{32.88}  & 32.74 \\ [-0.3ex]
			%\hline	
			 PSNR & 20.17 &  \textcolor{red}{29.13} & 28.94 & \textcolor{blue}{28.98} \\ [-0.3ex]
             SSIM & 0.6556 & \textcolor{red}{0.9232} & 0.9203 & \textcolor{blue}{0.9219} \\ 
			%\hline	
			 \textbf{Top-1 Acc} & 11.99 & 56.86 & \textcolor{blue}{57.87} &  \textcolor{red}{58.16} \\
             \textbf{Top-5 Acc} & 29.83 & 82.64 & \textcolor{blue}{83.65} & \textcolor{red}{83.70} \\ 
			\hline
%             \multirow{3}{*}{COCO 2014} %& PSNR & 24.94 & 32.58 & \textcolor{red}{32.96} & 32.08 & \textcolor{blue}{32.88}  & 32.74 \\ [-0.3ex]
% 			%\hline	
% 			& PSNR & 20.1719 &   32.7031&  32.4818& 32.6015 \\ [-0.3ex]
%             & SSIM & 0.3233  &  0.9137&0.9095 &0.9108  \\ 
% 			%\hline	
% 			\multirow{1}{*}{detection} 
%             & mAP &0.4401 & 0.5296 & 0.5268 &0.5330  \\

% 			\hline
		\end{tabular}	
	\end{center}
	\vspace{-2mm}
%	\vspace{-5mm}
\end{table}

\begin{table}%[th!]
	\begin{center}
%		\fontsize{8}{10pt}\selectfont
		\caption{Detection results after denoising noisy MS COCO images.}
		\label{table:task_transfer_detection}		
		\vspace{-2mm}
		\begin{tabular}{|@{\hskip 1mm}c@{\hskip 1mm}|@{\hskip 1mm}c@{\hskip 1mm}|c|@{\hskip 1mm}c@{\hskip 1mm}|@{\hskip 1mm}c@{\hskip 1mm}|@{\hskip 1mm}c@{\hskip 1mm}|}		% 5 columns
			\hline
			\multicolumn{2}{|c|}{} & noisy & CDnCNN-B & S-SAID & U-SAID \\   [-0.3ex] % reduce the space between two rows
			\hline 
			\hline
			\multirow{3}{*}{$\sigma$ = 15} 
			%\hline	
			& PSNR & 24.61 &  \textcolor{red}{35.14} & 34.92 & \textcolor{blue}{35.01}  \\ [-0.3ex]
            & SSIM &0.4796  & \textcolor{red}{0.9440} &  0.9410& \textcolor{blue}{0.9411}\\ 
            & \textbf{mAP} & 0.5110& \textcolor{blue}{0.5573} &0.5565 & \textcolor{red}{0.5590} \\
			\hline
            \multirow{3}{*}{$\sigma$ = 25} 
			%\hline	
			& PSNR & 20.17 &   \textcolor{red}{32.70}&  32.48 & \textcolor{blue}{32.60} \\ [-0.3ex]
            & SSIM & 0.3233  &  \textcolor{red}{0.9137}&0.9095 & \textcolor{blue}{0.9108}  \\ 
            & \textbf{mAP} &0.4401 & \textcolor{blue}{0.5296} & 0.5268 &\textcolor{red}{0.5330}  \\
			\hline
            \multirow{3}{*}{$\sigma$ = 35} 
			%\hline	
			& PSNR & 17.25 &  \textcolor{red}{31.12} &30.89 &  \textcolor{blue}{31.02}\\ [-0.3ex]
            & SSIM & 0.2383 & \textcolor{red}{0.8861} &  0.8803& \textcolor{blue}{0.8821}\\ 
            & \textbf{mAP} & 0.3663&\textcolor{blue}{0.5023}  & 0.4972 & \textcolor{red}{0.5056} \\
			\hline
		\end{tabular}	
	\end{center}
	\vspace{-3mm}
%	\vspace{-5mm}
\end{table}

For detection, We choose the MS COCO benchmark \cite{lin2014microsoft}, and add $\sigma$ = 15, 25, 35 noise to its validation set. We evaluate three denoisers in the same way as for the classification experiment, using a pre-trained YOLOv3 detection model \cite{redmon2018yolov3}. Table \ref{table:task_transfer_detection} shows consistent observations as above: U-SAID always leads to the largest improvements in the detection mean average prediction (mAP), and hence has the best task generalizablity among all. Another interesting observation is that S-SAID is not as competitive as CDnCNN-B for the detection task, which we leave for future work to explore.  

Both experiments show that the high-level semantics of different tasks are highly transferable for U-SAID, in terms of low-level vision tasks, as in line with \cite{liu2017image}.

\subsection{Statistical Significance Study of U-SAID's Improvement}

\begin{table}[th!]%[th!]
	\begin{center}
%		\fontsize{8}{10pt}\selectfont
% 		\caption{Performance and variance on three different tasks}
% 		\label{table:variance}		
% 		\vspace{-2mm}
		\begin{tabular}{|@{\hskip 1mm}c@{\hskip 1mm}|@{\hskip 1mm}c@{\hskip 1mm}|c|@{\hskip 1mm}c@{\hskip 1mm}|}	
			\hline
			& CDnCNN-B & S-SAID & U-SAID
% 			\multicolumn{1}{|c|}{} & CDnCNN-B & S-SAID & U-SAID
			\\  
			
% 			[-0.3ex] % reduce the space between two rows
			
			\hline 
			\hline
			\multicolumn{4}{|c|}{PASCAL VOC Segmentation} \\
			\hline
			\multirow{1}{*}
			%\hline	
			mIOU & 39.46\%  &  \textcolor{blue}{40.19\%}  &  \textcolor{red}{40.35\%}  \\ 
            Variance & 3.30E-6 &3.98E-6 & 3.15E-6 \\ 
			\hline
            \multicolumn{4}{|c|}{Cross-set Kodak Denoising}\\ \hline
            \multirow{1}{*}{ } 
			%\hline	
			NIQE & 2.87 & \textcolor{red}{2.60} & \textcolor{blue}{2.62} \\
            Variance & 1.74E-4 & 1.78E-4 &  6.00E-4  \\ 
			\hline
			\multicolumn{4}{|c|}{Cross-task CIFAR-100 classification}\\ \hline
            \multirow{1}{*}{} 
			%\hline	
			top-1 Accuracy & 56.89\%  & \textcolor{blue}{57.82\%}  & \textcolor{red}{58.47\%}  \\ 
            top-1 Variance & 0.03 & 0.06 &  0.02\\ 
            top-5 Accuracy & 82.89\%  & \textcolor{blue}{83.57\%}  & \textcolor{red}{83.91\%}  \\ 
            top-5 Variance & 0.02 & 0.05 &  0.06\\ 
            
            % & \textbf{mAP} & 0.3663&\textcolor{blue}{0.5023}  & 0.4972 & \textcolor{red}{0.5056} \\
			\hline
		\end{tabular}
	\caption{Performance and variance on three different tasks}
	\label{table:variance}
	\end{center}
% 	\vspace{-2mm}
	\vspace{-5mm}
\end{table}

How consistent and statistically meaningful is U-SAID's performance advantage? To answer this, we report the detailed statistics: (1) the $p$-values of the denoising quality improvement over different testing images; and (2) the variance of the performance improvements with different simulated noise patterns, for three representative experiments: PASCAL VOC segmentation (Table \ref{table:seg}), cross-set KODAK denoising (Table \ref{table:kodak_denoising}), and cross-task CIFAR-100 classification (Table \ref{table:task_transfer}). For each test, we simulated i.i.d. random Gaussian noise ($\sigma = 25$) for each image ten times, and repeat the experiments on them accordingly. Experiment results are shown in Table \ref{table:variance}.

In the PASCAL VOC segmentation experiment, we performance hypothesis tests to check if U-SAID leads to better segmentation results than CDnCNN-B. Being 95\% confident, we obtained $p$-value $= 1.7305E-9$, which demonstrates the statistical significance of improvement. On the other hand, U-SAID and S-SAID’s results do not show significant difference with $p$-value $= 0.0744 > 0.05$. Without using  any segmentation ground truth, our method achieved \textbf{statistically similar results} to S-SAID, even under a disadvantageous setting.

For the cross-set Kodak denoising experiment, the NIQE of U-SAID is statistically significantly better than that of CDnCNN-B, with $p$-value $= 2.6638E-16$. Similarly, S-SAID is better than U-SAID in NIQE with $p$-value $= 6.7845E-3$. 
%It solidifies our conclusion in paper. 

In CIFAR-100 experiment, for top-1 accuracy, U-SAID yields mean accuracy of 58.47\%, which is significantly higher than DnCNN, which has mean = 56.89\%, with $p$-value = 3.6147E-14. U-SAID has also higher accuracy than S-SAID (mean = 57.82\%) with $p$-value = 1.3486E-6. Similarly for top-5, U-SAID's performance ( 83.91\%) is statistically significant better than DnCNN (82.89\%), and S-SAID (83.57\%), with $p$-values of 1.3982E-9 and 4.3994E-3, respectively.
% \vspace{0.5cm}

\section{Conclusion}

This paper proposes a segmentation-aware image denoising model that requires no ground-truth segmentation map for training. The proposed U-SAID model leads to comparable performance with its supervised counterpart, in terms of both low-level (denoising) and high-level (segmentation) vision metrics, when trained on and applied to the same noisy dataset (without utilizing extra segmentation information as the latter has to). Furthermore, U-SAID shows remarkable generalizablity to unseen data, semantics, and high-level tasks, all of which endorse it to be a highly robust, effective and general-purpose denoising option.

%%%%%%%%%%%%%%%%%%%%%%%%%%%%%%%%%%%%%%%%%%%%%%%%%%%%%%%%%%%%%%%%%%%

% or


% \appendices

% % you can choose not to have a title for an appendix
% % if you want by leaving the argument blank
% \section{}

% % use section* for acknowledgment
% \section*{Acknowledgment}


\ifCLASSOPTIONcaptionsoff
  \newpage
\fi



%\bibliographystyle{IEEEtran}

\bibliographystyle{IEEEtran}
\bibliography{egbib}



% \end{thebibliography}


% \end{IEEEbiography}


% % insert where needed to balance the two columns on the last page with
% %\newpage



%\vfill


\end{document}